# Research on Medical Named Entity Identification Based On Prompt-Biomrc Model and Its Application in Intelligent Consultation System

Jinzhu Yang[✉]

NLP Applied Scientist, 525 Washington Blvd Suite 300, Jersey City, 07310, NJ, USA
[✉]Corresponding author: jy3024@columbia.edu

**Abstract.** This study is dedicated to exploring the application of prompt learning methods to advance Named Entity Recognition (NER) within the medical domain. In recent years, the emergence of large-scale models has driven significant progress in NER tasks, particularly with the introduction of the BioBERT language model, which has greatly enhanced NER capabilities in medical texts. Our research introduces the Prompt-bioMRC model, which integrates both hard template and soft prompt designs aimed at refining the precision and efficiency of medical entity recognition. Through extensive experimentation across diverse medical datasets, our findings consistently demonstrate that our approach surpasses traditional models. This enhancement not only validates the efficacy of our methodology but also highlights its potential to provide reliable technological support for applications like intelligent diagnosis systems. By leveraging advanced NER techniques, this study contributes to advancing automated medical data processing, facilitating more accurate medical information extraction, and supporting efficient healthcare decision-making processes.

## 1. Introduction

In the dynamic field of medical research and data analysis, there is an escalating need to efficiently extract and leverage information from extensive medical literature and diverse healthcare data. The conventional manual methods of annotating and processing these large datasets are not only costly and error-prone but also insufficient to meet the growing demands. Named Entity Recognition (NER) has emerged as a crucial subtask in this landscape, focusing on swiftly and accurately identifying specific entities such as diseases, medications, symptoms, and biological elements like DNA or proteins from unstructured medical texts. This capability not only accelerates entity identification processes and alleviates manual workload but also empowers healthcare professionals to make informed decisions and conduct research more effectively. The integration of advanced NER technologies demonstrated by the innovative Prompt-bioMRC model discussed in this paper promises to significantly improve the accuracy and efficiency of entity recognition in complex medical data environments. By utilizing a pre-defined language template or keyword-guided neural network's cue learning approach, the researchers aim to optimize NER performance for the multifaceted challenges inherent in processing diverse and complex medical texts. These advances not only aim to overcome the complexity of medical text data, which often spans multiple languages and domains, but also have a transformational impact in advancing healthcare delivery and biomedical innovation. By enhancing the automated data processing capabilities of this approach, it will help to discover new medical knowledge, support the development of personalized medicine strategies, and enhance the effectiveness of disease diagnosis, treatment, and prevention methods. As such, these advances are critical to the advancement of medical practice and biomedical research, helping to advance the evolution of more efficient healthcare systems and the exploration of personalized medicine and treatment strategies.

## 2. Related Research

### 2.1. Named Entity Recognition

Named Entity Recognition (NER) is a fundamental task in natural language processing (NLP) that involves identifying and classifying named entities within text into predefined categories such as names of persons, organizations, locations, dates, numerical values, and other types of entities. R. Hamad and colleagues introduced a Named Entity Recognition (NER) model designed to identify entities such as diseases, diagnoses, treatments, and symptoms. To achieve this, they utilized a dataset consisting of 27 authentic medical documents and performed classification tasks using a Support Vector Machine (SVM) based approach[1]. Y. Lu and collaborators utilized the Chinese electronic medical records from CCKS2019 as their experimental data source[2].They proposed an integrated model enhanced with knowledge graph (KG) embeddings and applied this model to tackle specific Chinese Clinical Named Entity Recognition (CNER) tasks. In their study, P. Chen and colleagues focused on enhancing the representation of electronic medical record texts by extracting local features and multi-level sequential interaction information[3].They proposed a hybrid neural network model, termed MC-BERT integrated with BiLSTM, CNN, Multi-Head Attention (MHA), and Conditional Random Field (CRF). In their paper, T. Yang and collaborators introduced a hybrid neural network model for Named Entity Recognition in medical texts[4],alongside proposing a novel multivariate convolutional decoding approach.

### 2.2. Tips for learning

Learning is a lifelong journey that requires dedication, persistence, and effective strategies to maximize comprehension and retention. TR Gadekallu and colleagues highlighted the unprecedented feasibility of remote patient monitoring enabled by advancements in Internet of Medical Things (IoMT) and wearable devices[5]. M. Li and colleagues introduced a novel approach called Weighted Prototype Contrastive Learning for medical few-shot Named Entity Recognition (W-PROCER)[6].Their method focuses on developing a framework based on prototype-based contraction loss and weighted networks. VK Vo-Ho and collaborators detailed various Neural

Architecture Search (NAS) methods in medical imaging, highlighting their applications in tasks[7] such as classification, segmentation, detection, and reconstruction. Furthermore, they elucidated the application of meta-learning within NAS for tasks involving few-shot learning and multi-tasking.

### 2.3. MRC model

The MRC (Machine Reading Comprehension) model is a model based on machine learning and natural language processing techniques, designed to enable computers to understand and answer questions in text. X. Du and collaborators proposed a multi-task learning and multi-strategy approach based on Machine Reading Comprehension (MRC)[8]. Experimental validation on the nested NER corpus CMeEE and the flat NER corpus CCKS2019 demonstrated the effectiveness of their MRC model leveraging multi-task learning and diverse strategies. C. Peng and colleagues employed a unified text-to-text learning framework based on generative large language models (LLM) to address key clinical natural language processing (NLP) tasks[9]. The proposed approach leveraged a unified generative LLM to achieve top performance in five out of seven major clinical NLP tasks. In their publication, Z. Zhang and colleagues proposed a novel model termed Coordinated Mobile and Residual Transformer UNet (MRC-TransUNet)[10], which integrates the strengths of both transformer and UNet architectures.

## 3. Method

### 3.1. Modeling framework

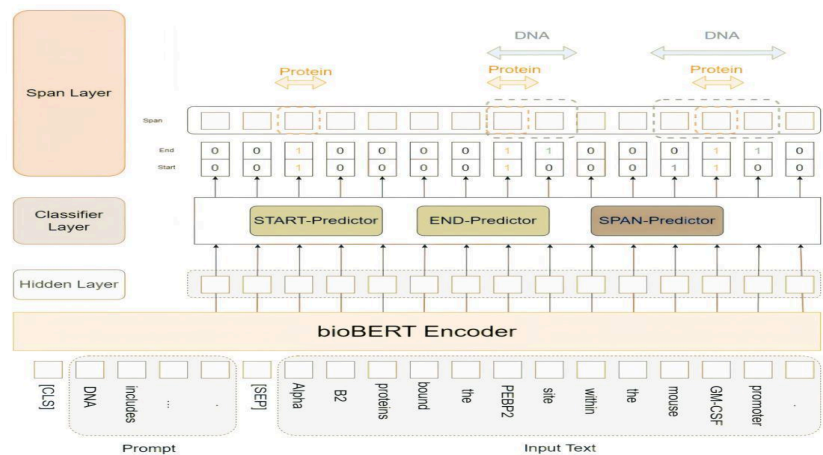

**Figure 1.** Model architecture of the Prompt-bioMRC

This model framework is designed to address the task of identifying entities within complex input sequences $X = \{x_1, x_2, ..., x_n\}$, where n signifies the length of the sequence. The primary objective is to assign labels $y \in Y$ to each entity, representing diverse categories such as individuals (PER), geographical locations (LOC), pharmaceutical products (Drug), and medical conditions (Disease). The data preprocessing pipeline involves multiple intricate stages. Initially, the labeled-style dataset for named entity recognition undergoes transformation into triplets structured as (PROMPT, ANCHOR, CONTEXT). Each entity label $y \in Y$ is associated with a prompt $P_y = \{p_1, p_2, ..., p_m\}$, where m denotes the length of the prompt. Annotated entities $x_{start,end} = \{x_{start}, x_{start+1}, ..., x_{end-1}, x_{end}\}$ represent substrings within $x$, delineated by indices start and end.

Through the tailored design of prompts $P_y$ for individual labels $y$, the resulting triplet (PROMPT, ANCHOR_START, ANCHOR_END, X) is structured to encompass sequences where indices start and end demarcate a continuous span of labels. As shown in Figure 1.

### 3.2. Main network

The core of our study's network architecture involves applying the Machine Reading Comprehension (MRC) framework to extract specific spans of text $x_{start,end}$ corresponding to entity type $y$ from the input sequence $X$. BioBERT serves as the foundational model, integrated with prompts $P_y$ concatenated alongside the textual input $X$. This concatenation results in a unified string format $\{[CLS], q_1, q_2, ..., q_m, [SEP], x_1, x_2, ...x_n\}$, incorporating markers [CLS] and [SEP] as defined in models similar to BERT. Following this, BioBERT processes the concatenated string to produce a contextual embedding matrix $E \in R^{n \times d}$, where d represents the vector dimensionality of BioBERT's final layer, encapsulating the representation of the original text passage $X$. As shown in Figure 2.

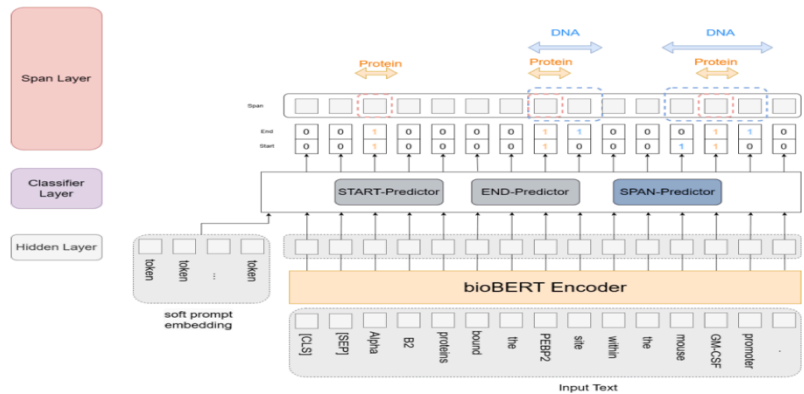

**Figure 2.** Model architecture of the soft-Prompt bioMRC

### 3.3. Experimental desig

In this section, comprehensive experimental analyses are conducted utilizing the CADEC dataset to evaluate a range of pretrained models beyond the foundational BERT-base architecture. Specifically, bioBERT, bio_clinical_BERT, and bio_clinical_medical_BERT, each pretrained on distinct corpora, undergo rigorous comparative evaluations aimed at discerning the most effective model configuration.

Conducted with meticulous care on the original Machine Reading Comprehension (MRC) task dataset, deliberately excluding the integration of prompts to maintain a rigorous baseline evaluation framework, the experiments meticulously document elaborate specifications of the experimental hyperparameters to ensure the transparency and reproducibility of the findings, as depicted in Table 1.

**Table 1.** Hyperparameters Settings of Pre-trained Model Layers

| Parameter | Value | Parameter | Value |
| --- | --- | --- | --- |
| Max num_epochs | 30 | hidden_dropout_prob | 0.1 |
| Layer_norm_eps | 1e-12 | hidden_size | 768/1024 |
| Vocab_size | 28996 | Num_attention_heads | 12/16 |
| Max_position_embeddings | 512 | Num_hidden_layers | 12/24 |

### 3.4. Soft-Prompt

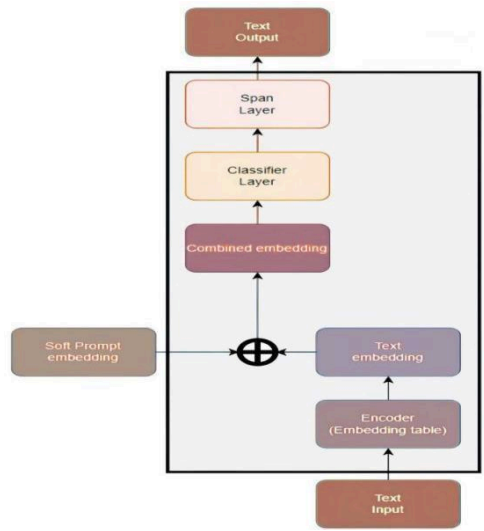

**Figure 3.** Schematic diagram of the design model for soft-Prompt

In developing the soft-Prompt model, the focus lies on integrating a dynamic prompting mechanism tailored to enhance task-specific performance on datasets while maintaining overall operational efficiency. Unlike rigid templates that necessitate manual construction, soft prompts leverage inherent learning mechanisms within the model to autonomously optimize performance. However, despite meticulous design efforts for soft prompts, there are instances where their effectiveness in model calibration may not surpass that achieved by manually designed hard templates. For example, experimental findings illustrate cases where the performance gap between the frozen GPT-3 model, equipped with 175 billion parameters, and a finely tuned T5 model, which operates with significantly fewer parameters—approximately 800 times less—was observed to be five percentage points. As shown in Figure 3.

## 4.    Results and discussion

In examining the experimental data presented in Figure 4, it becomes evident that BioBERT stands out as the top performer among all the evaluated pre-trained models. The analysis of the content of the pre-training datasets associated with each model category leads us to attribute BioBERT's superior performance to its alignment with the specific characteristics of the dataset employed in this study. In medical entity recognition tasks, precision is typically regarded as the primary metric, focusing on the model's ability to accurately determine the identities of designated entities. The objective is to maximize the count of accurately identified entities within the set predicted by the model. Within the scope of the medical domain tasks investigated in this research, BioBERT achieved a precision rate of 64.85%, reflecting an improvement of over two percentage points compared to the baseline model, BERT-large. Additionally, BioBERT exhibited a slightly superior F1-Score relative to the baseline model, underscoring its robust performance within the framework of the Prompt-bioMRC model and the chosen pre-training strategy. These findings underscore the effective utilization of selected pre-training models in the context of this study.

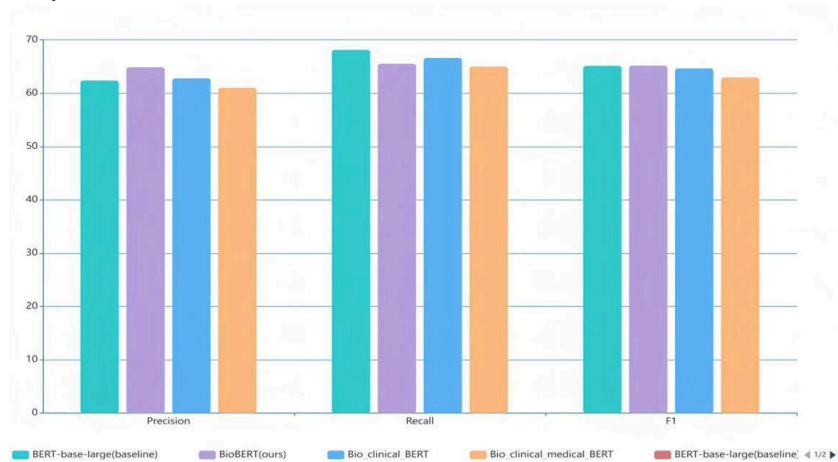

**Figure 4.** Comparison Experiments of Pre-trained Models

Figure 5 displays the assessment outcomes of our model proposed in this study on the GENIA dataset, juxtaposed with comparative analyses against other methodologies. Our research reveals a substantial rise in F1 score by nearly 5 percentage points compared to the Hyper-Graph model, showcasing varied improvements relative to models like ARN, Path-BERT, DYGIE, and Seq2seq-BERT. Furthermore, in contrast to the baseline MRCNER model, our approach demonstrates an enhanced F1 score by 0.86 percentage points.

The Prompt bioMRC model performed well on the GENIA dataset, surpassing the benchmarks set by other advanced models, highlighting its advantages in accuracy, recall, and F1 score metrics. This achievement highlights the advantages of our proposed method and provides strong evidence to support the effectiveness of prompt based ensemble techniques in medical named entity recognition (NER) tasks fully elaborated in this study. By verifying the feasibility and practicality of an instant learning paradigm tailored to the inherent complexity and nuances in medical texts, our study positions the real-time biological mRC model as an important advancement in this field.

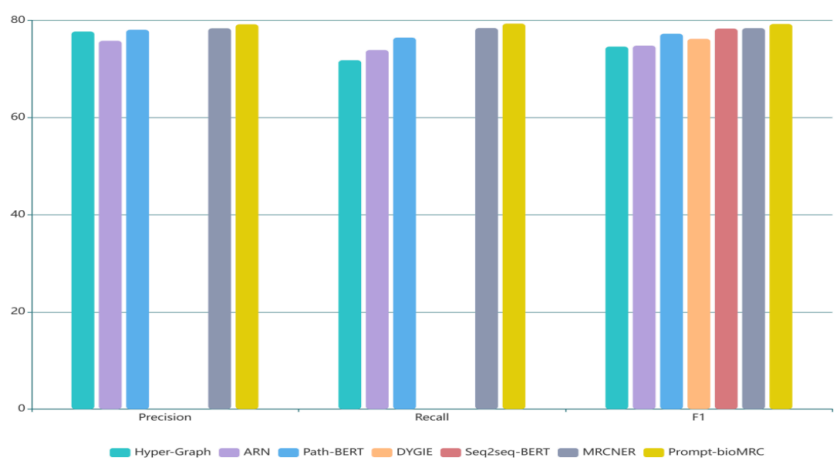

**Figure 5.** Comparison Experiments of Prompt-bioMRC and Previous Model

## 5. Conclusion

In the field of Named Entity Recognition (NER) tasks in medicine, researchers mainly rely on advanced pre trained models, such as BioBERT launched by Lee et al. in 2020, and professional developments such as UmlsBERT specifically designed for medical applications. However, there is still a difficulty that current medical NER research methods are often considered lagging behind the forefront advances in the broader field of natural language processing. This lag is particularly evident in the absorption and integration of the latest technologies such as rapid learning. In addition, the field also faces serious obstacles such as limited availability of annotated data, significant differences between clinical terms, and the non standardized nature of medical texts, all of which result in unsatisfactory results for the universal language model.

Through carefully designed and executed comparative experiments, this study provides strong evidence to support the effectiveness of integrating rapid engineering into medical named entity recognition (NER) tasks. Using well-designed hard and soft prompts can significantly improve model performance. Our research has found that using pre trained language models specifically designed for the medical field also has significant advantages in optimizing the results of specific tasks. To address the enduring challenges in medical NER tasks, this research introduces a novel Prompt-based approach tailored explicitly for medical NER, exploring its efficacy within this specialized domain. Looking forward, future research endeavors could delve deeper into refining medical NER tasks:

(1) Development of Soft-Prompts: Our exploration into Soft-Prompts remains in its preliminary stages, primarily focusing on initial model design, implementation, and initial exploratory analyses. Future inquiries may delve into more intricate Soft-Prompt mechanisms that integrate a broader array of parameters, potentially incorporating modular, plug-and-play frameworks that require further exploration.

(2) Enhancing the Universality of Hard Templates: The creation of prompts for medical datasets within this study was grounded in dataset-specific characteristics, involving significant manual labor and domain-specific expertise. Efforts to infuse more comprehensive background knowledge into hard templates could optimize their utility, thus maximizing model effectiveness across diverse medical contexts.

This examination of medical NER within this paper represents a modest contribution within the expansive realm of natural language processing. As research progresses and investigations deepen, we anticipate that boundaries within NER tasks and the broader field of natural language processing will expand in tandem with advancements in language modeling techniques and industrial applications..